\pdfoutput=1

\documentclass[11pt,a4paper]{article}
\usepackage[hyperref]{acl2020}
\usepackage{times}
\usepackage{latexsym}

\usepackage{microtype}

\aclfinalcopy
\def\aclpaperid{27}

\usepackage{algorithm}
\usepackage{algorithmic}
\usepackage{url}
\usepackage{graphicx}
\usepackage{booktabs}
\usepackage{multirow}
\usepackage{multicol}

\title{Learning Source Phrase Representations for Neural Machine Translation}

\author{
Hongfei Xu$^{1,2}$\ \ \ \ Josef van Genabith$^{1,2}$\thanks{\ \ \ \ Corresponding author.}\ \ \ \ Deyi Xiong$^3$\ \ \ \ Qiuhui Liu$^4$\ \ \ \ Jingyi Zhang$^2$\\
$^1$Saarland University / Saarland, Germany\\
$^2$German Research Center for Artificial Intelligence / Saarland, Germany\\
$^3$Tianjin University / Tianjin, China\\
$^4$China Mobile Online Services / Henan, China\\
hfxunlp@foxmail.com,
Josef.Van\_Genabith@dfki.de,
dyxiong@tju.edu.cn,\\
liuqhano@foxmail.com,
Jingyi.Zhang@dfki.de
}
\date{}

\begin{document}
\maketitle
\begin{abstract}
The Transformer translation model \cite{vaswani2017attention} based on a multi-head attention mechanism can be computed effectively in parallel and has significantly pushed forward the performance of Neural Machine Translation (NMT). Though intuitively the attentional network can connect distant words via shorter network paths than RNNs, empirical analysis demonstrates that it still has difficulty in fully capturing long-distance dependencies \cite{tang2018self}. Considering that modeling phrases instead of words has significantly improved the Statistical Machine Translation (SMT) approach through the use of larger translation blocks (``phrases'') and its reordering ability, modeling NMT at phrase level is an intuitive proposal to help the model capture long-distance relationships. In this paper, we first propose an attentive phrase representation generation mechanism which is able to generate phrase representations from corresponding token representations. In addition, we incorporate the generated phrase representations into the Transformer translation model to enhance its ability to capture long-distance relationships. In our experiments, we obtain significant improvements on the WMT 14 English-German and English-French tasks on top of the strong Transformer baseline, which shows the effectiveness of our approach. Our approach helps Transformer Base models perform at the level of Transformer Big models, and even significantly better for long sentences, but with substantially fewer parameters and training steps. The fact that phrase representations help even in the big setting further supports our conjecture that they make a valuable contribution to long-distance relations.
\end{abstract}

\section{Introduction}

NMT is a new approach to machine translation that has achieved great success in the
last a few years \cite{sutskever2014sequence,bahdanau2014neural,gehring2017convolutional,vaswani2017attention}. Compared to plain SMT \cite{brown1993mathematics,koehn2003statistical,chiang2005hierarchical}, a neural language model decoder \cite{sutskever2014sequence} is better at long-distance re-ordering, and attention mechanisms \cite{bahdanau2014neural,vaswani2017attention} have been proven effective in modeling long-distance dependencies, while these two issues were both challenging for SMT.

The Transformer \cite{vaswani2017attention}, which has outperformed previous RNN/CNN based translation models \cite{bahdanau2014neural,gehring2017convolutional}, is based on multi-layer multi-head attention networks and can be trained in parallel very efficiently. Though attentional networks can connect distant words via shorter network paths than RNNs, empirical results show that its ability in capturing long-range dependencies does not significantly outperform RNNs, and it is still a problem for the Transformer to fully model long-distance dependencies \cite{tang2018self}.

Using phrases instead of words enables conventional SMT to condition on a wider range of context, and results in better performance in re-ordering and modeling long-distance dependencies. It is intuitive to let the NMT model additionally condition on phrase level representations to capture long-distance dependencies better, but there are two main issues which prevent NMT from directly using phrases:
\begin{itemize}
\item There are more phrases than tokens, and the phrase table is much larger than the word vocabulary, which is not affordable for NMT;
\item Distribution over phrases is much sparser than that over words, which may lead to data sparsity and hurt the performance of NMT.
\end{itemize}

Instead of using phrases directly in NMT, in this work, we address the issues above with the following contributions:

\begin{itemize}
\item To address the large phrase table issue, we propose an attentive feature extraction model and generate phrase representation based on token representations. Our model first summarizes the representation of a given token sequence with mean or max-over-time pooling, then computes the attention weight of each token based on the token representation and the summarized representation, and generates the phrase representation by a weighted combination of token representations;
\item To help the Transformer translation model better model long-distance dependencies, we let both encoder layers and decoder layers of the Transformer attend the phrase representation sequence which is shorter than the token sequence, in addition to the original token representation. Since the phrase representations are produced and attended at each encoder layer, the encoding of each layer is also enhanced with phrase-level attention computation;
\item To the best of our knowledge, our work is the first to model phrase representations and incorporating them into the Transformer.
\end{itemize}

Our approach empirically brings about significant and consistent improvements over the strong Transformer model (both base and big settings). We conducted experiments on the WMT 14 English-German and English-French news translation task, and obtained $+1.29$ and $+1.37$ BLEU improvements respectively on top of the strong Transformer Base baseline, which demonstrates the effectiveness of our approach. Our approach helps Transformer Base models perform at the level of Transformer Big models, and even significantly better for long sentences, but with substantially fewer parameters and training steps. It also shows effectiveness with the Transformer Big setting. We also conducted length analysis with our approach, and the results show how our approach improves long-distance dependency capturing, which supports our conjecture that phrase representation sequences can help the model capture long-distance relations better.

\section{Background and Related Work}

In this section, we first review previous work which utilizes phrases in recurrent sequence-to-sequence models, then give a brief introduction to the stronger Transformer translation model that our work is based on.

\subsection{Utilizing Phrases in RNN-based NMT}

Most previous work focuses on utilizing phrases from SMT in NMT to address its coverage \cite{tu2016modeling} problem.

\citet{dahlmann2017neural} suggested that SMT usually performs better in translating rare words and profits from using phrasal translations, even though NMT achieves better overall translation quality. They introduced a hybrid search algorithm for attention-based NMT which extended the beam search of NMT with phrase translations from SMT. \citet{xing2017neural} proposed that while NMT generally produces fluent but often inadequate translations, SMT yields adequate translations though less fluent. They incorporate SMT into NMT through utilizing recommendations from SMT in each decoding step of NMT to address the coverage issue and the unknown word issue of NMT. \citet{wang2017translating} suggested that phrases play a vital role in machine translation, and proposed to translate phrases in NMT by integrating target phrases from an SMT system with a phrase memory given that it is hard to integrate phrases into NMT which reads and generates sentences in a token-by-token way. The phrase memory is provided by the SMT model which dynamically picks relevant phrases with the partial translation from the NMT decoder in each decoding step.

\subsection{The Transformer Translation Model}

\begin{figure}[t]
\centering
\includegraphics[width=1.0\columnwidth]{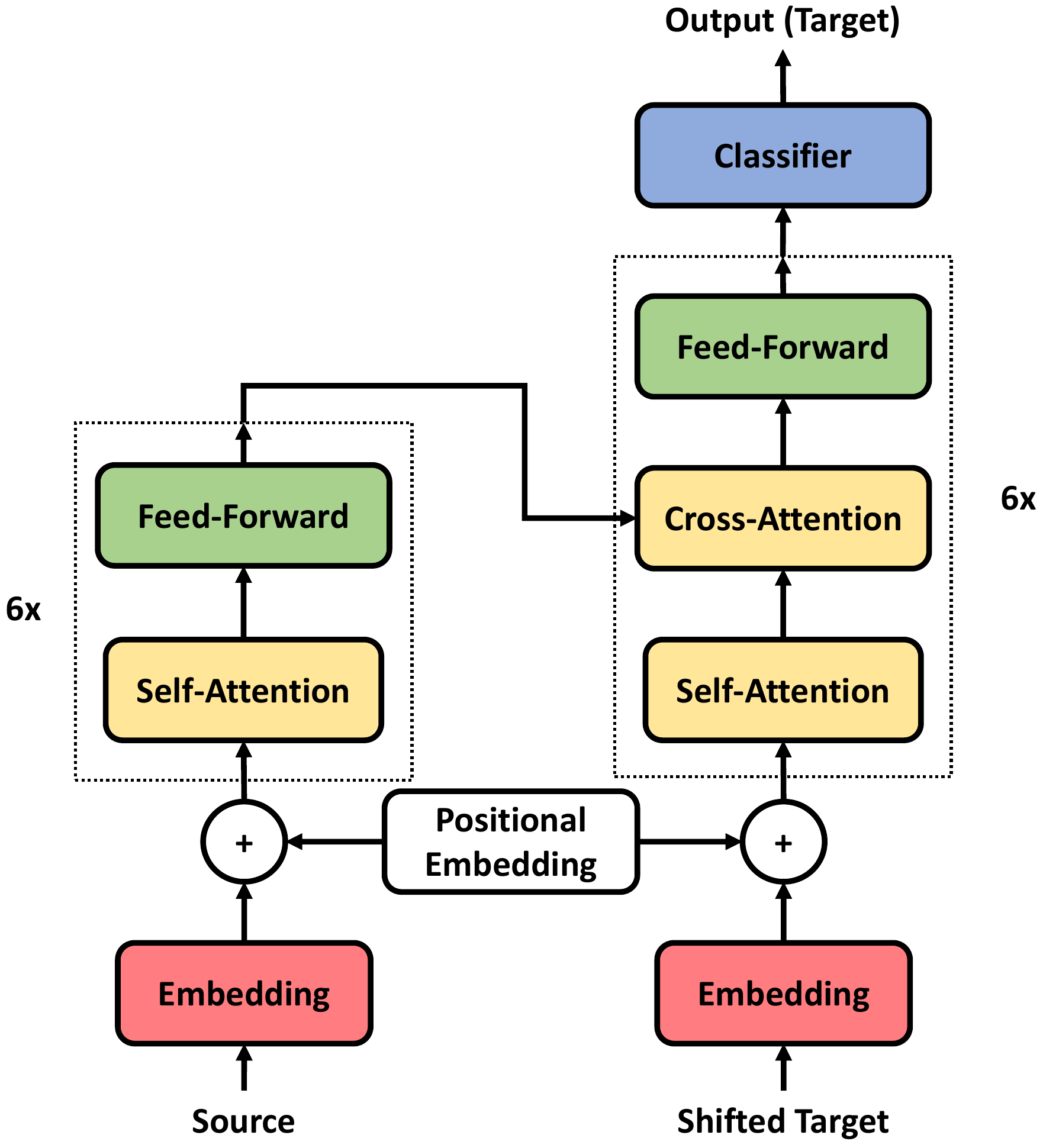}
\caption{The Transformer Translation Model. Residual connection and Layer normalization are omitted for simplicity.}
\label{Transformer}
\end{figure}

Our research is based on the Transformer translation model \cite{vaswani2017attention} shown in Figure \ref{Transformer}, which significantly outperforms the previous recurrent sequence-to-sequence approach and can be efficiently computed in parallel.

The Transformer includes an encoder and a decoder. Both encoder and decoder are a stack of 6 layers. Besides the embedding matrix and positional embedding matrix in both encoder and decoder, the decoder also has a softmax classifier layer to produce translated tokens. The weights of the softmax classifier are normally tied to the target embedding matrix.

Both encoder layers and decoder layers make use of the multi-head attention mechanism. The multi-head attention mechanism calculates attention results of given queries on corresponding keys and values. It first projects queries, keys and values with $3$ independent linear transformations, then splits the transformed key, query and value embeddings into several chunks of $d_k$ dimension vectors, each chunk is called a head,\footnote{$d_k$ is $64$ for both the Transformer Base and the Transformer Big, and the numbers of heads for them are $8$ and $16$ respectively.} and scaled dot-product attention is independently applied in each head:

\begin{equation}
     Attn(Q,K,V) = {\mathop{\rm softmax}\nolimits} (\frac{{Q{K^T}}}{{\sqrt {{d_k}} }})V
\end{equation}

\noindent where $Q$, $K$ and $V$ stand for the query vectors, key vectors and value vectors. Finally, the network concatenates the outputs of all heads and transforms it into the target space with another linear layer. The self-attention network uses the query sequence also as the key sequence and the value sequence in computation, while the cross-attention feeds another vector sequence to attend as queries and values.

Comparing the computation of the attentional network with RNNs, it is obvious that the attention computation connects distant words with a shorter network path, and intuitively it should perform better in capturing long-distance dependencies. However, empirical results show that its ability in modeling long-range dependencies does not significantly outperform RNNs.

\subsection{Comparison with Previous Works}

Compared to previous works using RNN-based NMT \cite{wei2016neural,xing2017neural,wang2017translating,dahlmann2017neural}, our proposed approach is based on the Transformer model, with the following further important differences:

\begin{itemize}
\item Our approach aims to improve the long-distance dependency modeling ability of NMT instead of coverage \cite{tu2016modeling};
\item Our approach does not require to train an SMT system or to extract aligned phrase translation from the training corpus, which makes it efficient and avoids suffering from potential error propagation from the SMT system. The phrase representation learning model is a neural model, and is deeply integrated in the translation model, and the whole neural model is end-to-end trainable;
\item We iteratively and dynamically generate phrase representations with token vectors. Previous work does not use SMT phrases in this way.
\end{itemize}

In more recent work, \newcite{wang2019self} augment self attention with structural position representations to model the latent structure of the input sentence; \newcite{hao2019multi} propose multi-granularity self-attention which performs phrase-level attention with several attention heads.

\section{Transformer with Phrase Representation}

For the segmentation of phrases, given that N-gram phrases are effective for tensor libraries, we first try to cut a token sequence into a phrase sequence with a fixed phrase length which varies with the sequence length.\footnote{We implement this as: $ntok = max(min(8, seql / 6), 3)$, where $ntok$ and $seql$ stand for the number of tokens in each phrase and the length of a sentence respectively.} We pad the last phrase in case it does not have sufficient tokens, thus we can transform the whole sequence into a tensor.

The N-gram phrase segmentation is efficient and simple, and we suggest the drawbacks of such ``casual'' segmentation boundaries can be alleviated with self-attention computation across the whole sequence and the attention mechanism applied in the generation of phrase representation which values tokens differently to a large extent, given that neural models have been proven good at learning competitively effective representations with gate or attention mechanism even without modeling linguistic structures \cite{cho2014learning,Hochreiter1997LSTM,vaswani2017attention,devlin2019bert}.

In our experiments we also explore phrases extracted from the Stanford Parser \cite{socher2013parsing} as as an alternative to our simple segmentation strategy. The maximum number of tokens allowed is consistent with the simple segmentation approach, and we try to use the tokens from the largest sub-tree that complies with the maximum token limitation or from several adjacent sub-trees of the same depth as a phrase for efficiency. Our algorithm to extract phrases from parse trees is shown in Algorithm \ref{alg:eppt}.

\begin{algorithm}[t]
\caption{Extracting Phrases from a Parse Tree. Input: A parse tree $T$, maximum tokens allowed in a phrase $n$; Output: Extracted phrase sequence $S$.}
\label{alg:eppt}
\begin{algorithmic}[1]
\WHILE{T is not empty}
\STATE Initialize a phrase sequence $p=[]$, maximum tokens allowed in this phrase $mt=n$;
\STATE Find the largest sub-tree $ST$ with $nst$ tokens ($nst < n$) and depth $dst$ from the right side of $T$;
\STATE Add the token sequence in $ST$ into $p$;
\STATE Remove $ST$ from $T$;
\WHILE{$mt>0$}
\STATE Find the adjacent sub-tree ${\mathop{\rm STA}\nolimits}$ of depth $dst$ with $nsta$ tokens from the right side of $T$;
\IF{${\mathop{\rm STA}\nolimits}$ exists and $nsta \le mt$}
    \STATE Insert the token sequence of ${\mathop{\rm STA}\nolimits}$ to the beginning of $p$;
    \STATE Remove ${\mathop{\rm STA}\nolimits}$ from $T$;
    \STATE $mt=mt-nsta$;
\ELSE
    \STATE Break;
\ENDIF
\ENDWHILE
\STATE Append $p$ to $S$;
\ENDWHILE
\STATE Reverse $S$;
\RETURN $S$
\end{algorithmic}
\end{algorithm}

To efficiently parallelize parser-based phrases of various length in a batch of data, we pad short phrases to the same length of the longest phrases in the batch of sentences, thus a batch of sequences of phrases can be saved into a tensor. But significantly more ``$<$pad$>$'' tokens will be introduced, and the model is slightly slower than the simple approach.

\begin{figure*}[t]
\centering
\includegraphics[width=1.5\columnwidth]{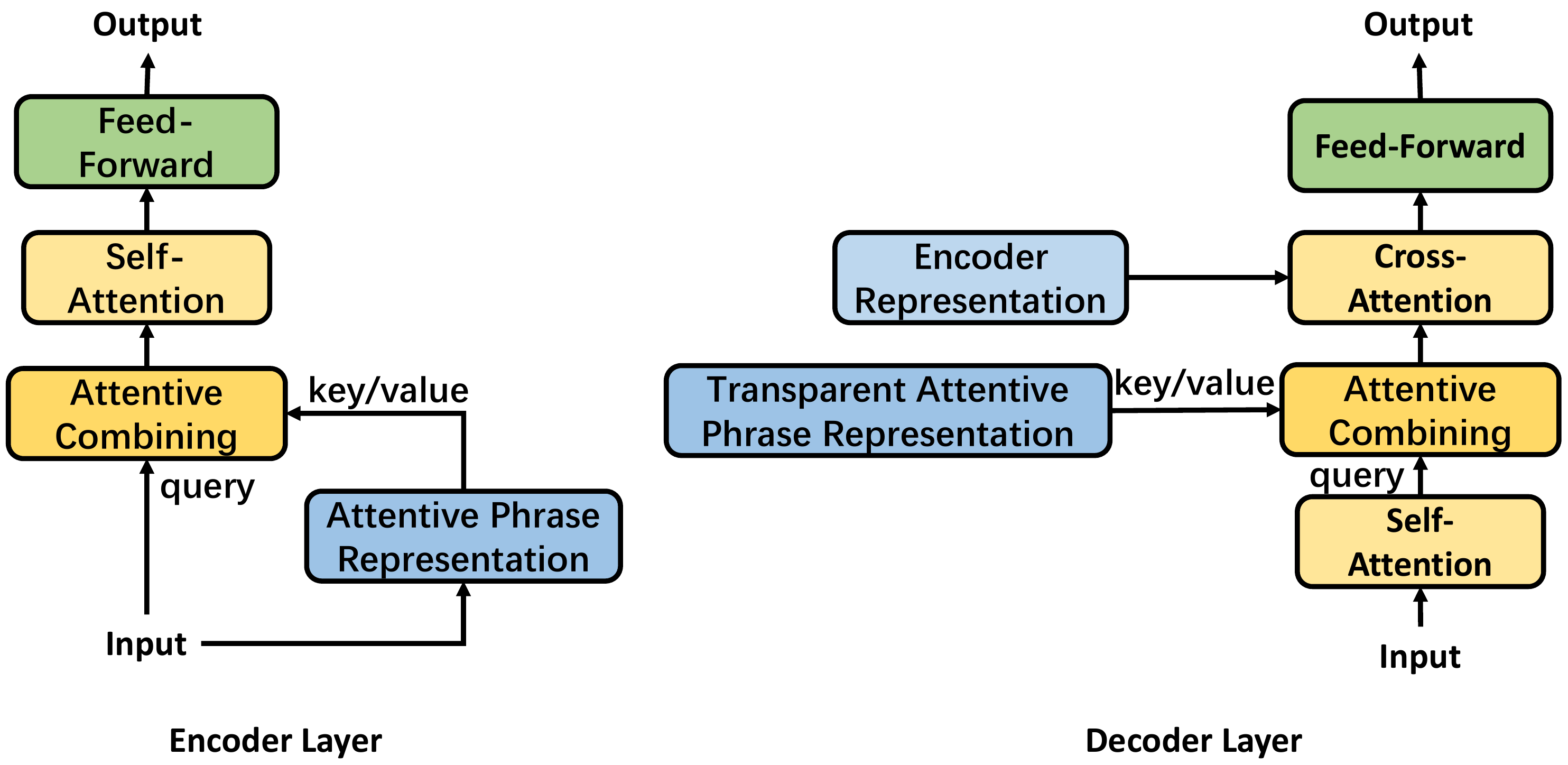}
\caption{The Encoder/Decoder Layer of the Transformer Model with Phrase Representation. Residual connection and Layer normalization are omitted for simplicity.}
\label{phrase}
\end{figure*}

\subsection{Attentive Phrase Representation Generation}

Merging several token vectors into one is very likely to incur information loss, and introducing an importance evaluation mechanism is better than treating tokens equally. To highlight the most important features in a segmented phrase chunk, we introduce an attentive phrase representation generation model to value tokens differently according to their importance in the phrase. The model first roughly extracts features from all tokens into a vector, then assigns a score to each token by comparing each token vector with the extracted feature vector, and produces the weighted accumulation of all token vectors according to their scores.

Phrase representations are generated in every encoder layer, for the $k_{th}$ encoder layer, we generate phrase representation $R_{e_{phrase}}^k$ from its input representation. Assume the phrase contains $m$ tokens $\{t_1, ... , t_m\}$, and $\{R_{e_{t_1}}^k, R_{e_{t_2}}^k, ... , R_{e_{t_m}}^k\}$ are the corresponding input vectors to the encoder layer, we first generate a summary representation by:

\begin{equation}
     {R_{e_{all}}^k} = F_{glance}({R_{e_{{t_1}}}^k},...,{R_{e_{{t_m}}}^k})
     \label{eqa:fsum}
\end{equation}

\noindent where $F_{glance}$ is a function to extract features of the vector sequence into a fixed-dimension vector; We explore both element-wise mean operation and max-over-time pooling operation in our work.

After the summarized representation is produced, we calculate a score for each token in the phrase, the score of the $i_{th}$ token $s_i^k$ is calculated as:

\begin{equation}
     {\rm{s}}_i^k = W_2^k\sigma (W_1^k[R_{{e_{{t_i}}}}^k|R_{{e_{all}}}^k] + b_1^k) + b_2^k
\end{equation}

\noindent where $\sigma$ is the sigmoid activation function, and ``$|$'' means concatenation of vectors. The rationale for designing this approach is further explained below.

Then we normalize the score vector to weights with the softmax function, and the probability of the $i_{th}$ token $p_i^k$ is:

\begin{equation}
    {\rm{p}}_i^k = \frac{{{e^{{\rm{s}}_i^k}}}}{{\sum\limits_{i = 1}^m {{e^{{\rm{s}}_i^k}}} }}
\end{equation}

Finally, the representation of the phrase in the $k_{th}$ encoder layer $R_{e_{phrase}}^k$ is generated by a weighted combination of all vectors:

\begin{equation}
     {R_{e_{phrase}}^k} = \sum\limits_{i = 1}^m {{p_i^k}{R_{e_{{t_i}}}^k}}
\end{equation}

The representation of the phrase sequence can be computed efficiently in parallel. Each encoder layer will produce a vector sequence as the phrase representation. We do not use the multi-head attention in the computation of the phrase-representation attention because of two reasons:

\begin{itemize}
\item The multi-head attention calculates weights through dot-product, we suggest that a 2-layer neural network might be more powerful at semantic level feature extraction, and it is less likely to be affected by positional embeddings which are likely to vote up adjacent vectors;
\item Though we employ a 2-layer neural network, it only has one linear transformation and a vector to calculate attention weights, which contains fewer parameters than the multi-head attention model that has 4 linear transformations.
\end{itemize}

Recent studies show that different encoder layers capture linguistic properties of different levels \cite{peters2018deep}, and aggregating layers is of profound value to better fuse semantic information \cite{shen2018dense,dou2018exploiting,wang2018multilayer,dou2019dynamic}. We assume that different decoder layers may value different levels of information i.e. the representation of different encoder layers differently, thus we weighted combined phrase representations from every encoder layer for each decoder layer with the Transparent Attention (TA) mechanism \cite{bapna2018training}. For the decoder layer $j$, the phrase representation ${R_{d_{phrase}}^j}$ fed into that layer is calculated by:

\begin{equation}
     R_{{d_{phrase}}}^j = \sum\limits_{i = 0}^d {w_i^jR_{{e_{phrase}}}^i}
\end{equation}

\noindent where $w_i^j$ are softmax normalized parameters trained jointly with the full model to learn the importance of encoder layers for the $j_{th}$ decoder layer. $d$ is the number of encoder layers, and $0$ corresponds to the embedding layer.

\subsection{Incorporating Phrase Representation into NMT}

After the phrase representation sequence for each encoder layer and decoder layer is calculated with the approach described above, we propose an attentive combination network to incorporate the phrase representation for each layer into the Transformer translation model to aid it modeling long-distance dependencies. The attentive combination network is inserted in each encoder layer and each decoder layer to bring in information from the phrase representation. The structures of the encoder layer and the decoder layer of the Transformer model with phrase representation are shown in Figure \ref{phrase}.

For an encoder layer, the new computation order is: cross-attention to phrases $\rightarrow$ self-attention over tokens $\rightarrow$ feed-forward neural network to process collected features, while for a decoder layer it is: self-attention over decoded tokens $\rightarrow$ cross-attention to source phrases $\rightarrow$ cross-attention to source tokens $\rightarrow$ feed-forward neural network to process collected features. Compared to the computation order of the standard Transformer, the new computation order performs additional attending at phrase level before attending source token representations at token level. We conjecture that attending at phrase level should be easier than at token level, and attention results at phrase level may aid the attention computation at the token-level.

For a given input sequence $x$ and a phrase vector sequence $R_{phrase}$, the attentive combination network first attends the phrase representation sequence and computes the attention output $out_{phrase}$ as follows:

\begin{equation}
     out_{phrase} = {\mathop{\rm Attn_{MH}}\nolimits}(x,{R_{phrase}})
\end{equation}

\noindent where ${\mathop{\rm Attn_{MH}}\nolimits}$ is a multi-head cross-attention network with $x$ as keys and $R_{phrase}$ as corresponding queries and values.

The attention result is then combined again with the original input sequence $x$ with a 2-layer neural network which aims to make up for potential information loss in the phrase representation with the original token representation:

\begin{equation}
     out = W_4\sigma (W_3[x|out_{phrase}] + b_3) + b_4
\end{equation}

We also employ a residual connection around the attentive combination layer, followed by layer normalization to stabilize the training.

Since the phrase representation is produced inside the Transformer model and utilized as the input of layers, and all related computations are differentiable, the attentive phrase representation model is simply trained as part of the whole model through backpropagation effectively.

\section{Experiments}

To compare with \citet{vaswani2017attention}, we conducted our experiments on the WMT 14 English to German and English to French news translation tasks.

\subsection{Settings}

We implemented our approaches based on the Neutron implementation \citep{xu2019neutron} of the Transformer translation model. We applied joint Byte-Pair Encoding (BPE) \cite{sennrich2015neural} with $32k$ merge operations on both data sets to address the unknown word problem. We only kept sentences with a maximum of $256$ subword tokens for training. Training sets were randomly shuffled in every training epoch. The concatenation of newstest 2012 and newstest 2013 was used for validation and newstest 2014 as test sets for both tasks.

The number of warm-up steps was set to $8k$, and each training batch contained at least $25k$ target tokens. Our experiments run on $2$ GTX 1080 Ti GPUs, and a large batch size was achieved through gradient accumulation. We used a dropout of $0.1$ for all experiments except for the Transformer Big on the En-De task which was $0.3$. The training steps for Transformer Base and Transformer Big were $100k$ and $300k$ respectively following \newcite{vaswani2017attention}. The other settings were the same as \cite{vaswani2017attention} except that we did not bind the embedding between the encoder and the decoder for efficiency.

We used a beam size of $4$ for decoding, and evaluated tokenized case-sensitive BLEU \footnote{\url{https://github.com/moses-smt/mosesdecoder/blob/master/scripts/generic/multi-bleu.perl}} with the averaged model of the last $5$ checkpoints for Transformer Base and $20$ checkpoints for Transformer Big saved with an interval of $1,500$ training steps \cite{vaswani2017attention}. We also conducted significance tests \cite{koehn2004statistical}.

\begin{table}[t]
  \centering
    \begin{tabular}{lll}
    \toprule
    Models & \multicolumn{1}{l}{En-De} & \multicolumn{1}{l}{En-Fr} \\
    \midrule
    Transformer Base  & 27.38  & 39.34 \\
    \multicolumn{1}{c}{+PR}   & \textbf{28.67}$^\dag$  & \textbf{40.71}$^\dag$ \\
    \midrule
    Transformer Big   & 28.49 & 41.36 \\
    \multicolumn{1}{c}{+PR}   & \textbf{29.60}$^\dag$ & \textbf{42.45}$^\dag$ \\
    \bottomrule
    \end{tabular}
  \caption{Results on WMT 14 En-De and En-Fr.}
  \label{tab:bleuall}
\end{table}

\begin{table*}[t]
	\centering
    \begin{tabular}{lrrrrr}
    \toprule
    \multicolumn{1}{c}{\multirow{2}[2]{*}{Models}} & \multicolumn{1}{c}{\multirow{2}[2]{*}{BLEU}} & \multicolumn{1}{c}{\multirow{2}[2]{*}{$\Delta$}} & \multicolumn{1}{c}{\multirow{2}[2]{*}{Para. (M)}} & \multicolumn{2}{c}{Time} \\
          &       &       &       & \multicolumn{1}{l}{Train} & \multicolumn{1}{l}{Decode} \\
    \midrule
    Transformer Base & 27.38  &       & 88.1  & \multicolumn{1}{l}{1.00x} & \multicolumn{1}{l}{1.00x} \\
    \midrule
    +Mean & 27.99  & 0.61  & \multirow{2}[2]{*}{129.0} & \multicolumn{1}{l}{1.64x} & \multicolumn{1}{l}{1.45x} \\
    +Max  & 28.13  & 0.75  &       & \multicolumn{1}{l}{1.60x} & \multicolumn{1}{l}{1.40x} \\
    \midrule
    +Max+Attn & 28.52  & 1.14  & \multirow{3}[2]{*}{173.0} & \multicolumn{1}{l}{1.74x} & \multicolumn{1}{l}{1.52x} \\
    +Max+Attn+TA & 28.67  & 1.29  &       & \multicolumn{1}{l}{1.75x} & \multicolumn{1}{l}{1.53x} \\
    +Max+Attn+TA+Parsing Phrase & \textbf{28.76} & \textbf{1.38} &       & \multicolumn{1}{l}{1.83x} & \multicolumn{1}{l}{1.60x} \\
    \midrule
    Transformer Big & 28.49  & 1.11  & 264.1  & \multicolumn{1}{l}{7.73x} & \multicolumn{1}{l}{2.68x} \\
    \bottomrule
    \end{tabular}
	\caption{Ablation Study. $\Delta$ indicates the BLEU improvements compared to the Transformer Base. Time represents the time consumption compared to the Transformer Base (in training and decoding). The Transformer Big consumes $3$ times training steps of the Transformer Base.}
	\label{tab:abl}
\end{table*}

\subsection{Main Results}

We applied our approach to both the Transformer Base setting and the Transformer Big setting, and conducted experiments on both tasks to validate the effectiveness of our approach. Since parsing a large training set (specifically, the En-Fr dataset) is slow, we did not use phrases from parse results in this experiment (reported in Table \ref{tab:bleuall}). Results are shown in Table \ref{tab:bleuall}. $\dag$ indicates $p<0.01$ compared to the baseline for the significance test.

Table \ref{tab:bleuall} shows that modeling phrase representation can bring consistent and significant improvements on both tasks, and benefit both the Transformer Base model and the stronger Transformer Big model. ``+PR'' is the Transformer with Phrase Representation, corresponding to the ``+Max+Attn+TA'' setting in Table \ref{tab:abl}.

The En-Fr task used a larger dataset ($\sim 36M$ sentence pairs) and achieved a higher baseline BLEU than the En-De task, we suggest significant improvements obtained by our approach on the En-Fr task with the Transformer Big supports the effectiveness of our approach in challenging settings.

\subsection{Ablation Study}

We also conducted a Transformer Base based ablation study on the WMT 14 En-De task to assess the influence of phrase representation, attention mechanism in phrase representation generation, transparent attention and phrases from parser output on performance. Results are shown in Table \ref{tab:abl}.

``+Mean'' and ``+Max'' are only using element-wise mean operation and max-over-time pooling to generate an initial rough phrase representation of a given token sequence. ``+Attn'' indicates generating phrase representations with our attentive approach, on top of the max-over-time pooling as $F_{glance}$ in Equation \ref{eqa:fsum}. ``+TA'' indicates use of the Transparent Attention mechanism to fuse information generated from every encoder layer for different decoder layers,\footnote{This only introduces an additional $7*6$ parameter matrix, which does not show significant influence in view of the amount of parameters.} otherwise only outputs of the last encoder layer are fed into all decoder layers. ``+Parse'' means using phrases extracted from parse results with Algorithm \ref{alg:eppt}.

Table \ref{tab:abl} shows that introducing phrase representation can significantly improve the strong Transformer Base baseline, even only with a simple element-wise mean operation over token representations brings about a $+0.61$ BLEU improvement ($p<0.01$). Summarizing representations with max-over-time pooling performs slightly better than with the element-wise mean operation. Our attentive phrase representation generation approach can bring further improvements over the max-over-time pooling approach. Though utilizing phrases from the parser can make use of linguistic knowledge and obtains most improvements, our simple and effective segmenting approach performs competitively, and we interpret these comparisons to show the positive effects of collapsing token sequences into shorter phrase sequences on the modeling of long-distance dependencies.

Though a significant amount of parameters are introduced for incorporating phrase representation into the Transformer model, our approach (``+Max+Attn+TA'') improved the performance of the Transformer Base model by $+1.29$ BLEU on the WMT 14 En-De news task, and the proposed Transformer model with phrase representation still performs competitively compared to the Transformer Big model with only about half the number of parameters and $1/3$ of the training steps. Thus, we suggest our improvements are not only because of introducing parameters, but also due to the modeling and utilization of phrase representation.

\begin{figure}[t]
\centering
\includegraphics[width=1.0\columnwidth]{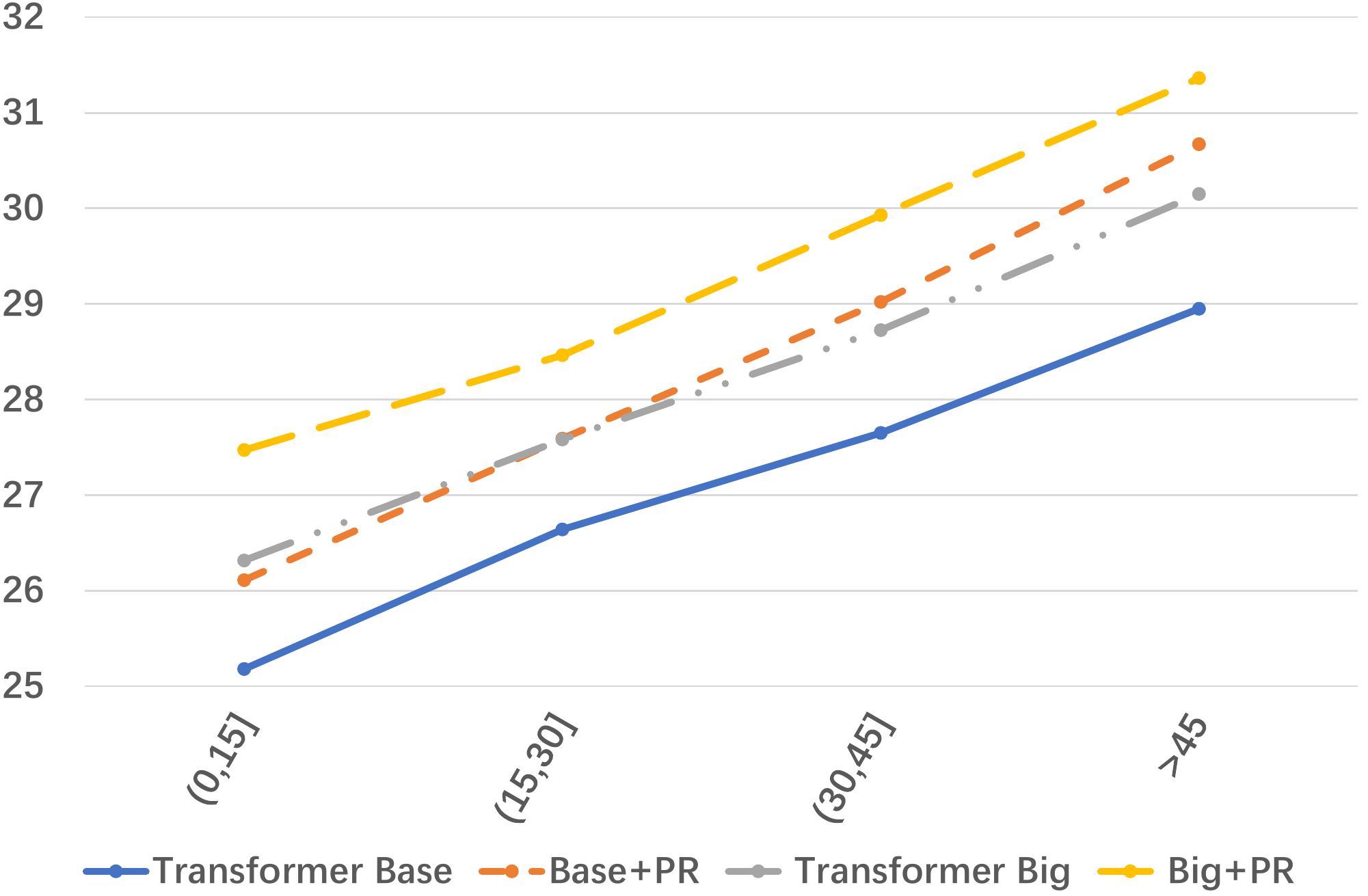}
\caption{BLEU scores with respect to various input sentence lengths.}
\label{length}
\end{figure}

\subsection{Length Analysis}

To analyze the effects of our phrase representation approach on performance with increasing input length, we conducted a length analysis on the news test set of the WMT 14 En-De task. Following \citet{bahdanau2014neural} and \citet{tu2016modeling}, we grouped sentences of similar lengths together and computed BLEU scores of Transformers and Transformers with phrase representations for each group. Results are shown in Figure \ref{length}.

Figure \ref{length} shows that our approach incorporating phrase representation into the Transformer significantly improves its performance in all length groups, and longer sentences show significantly more improvements than shorter sentences. In the Transformer Base setting, our approach improved the group with sentences of more than $45$ tokens by $+1.72$ BLEU, almost twice of the improvements for sentences with less than $15$ tokens which was $+0.93$ BLEU.

The effects of incorporating phrase representations into the Transformer is more significant especially when compared to the Transformer Big which has about twice the number of parameters than our approach and consumes $3$ times the training steps. According to \citet{tang2018self}, the number of attention heads in Transformers impacts their ability to capture long-distance dependencies, and specifically, many-headed multi-head attention is essential for modeling long-distance phenomena with only self-attention. The Transformer Big model with twice the number of heads in the multi-head attention network compared to those in the Transformer Base model, should be better at capturing long-distance dependencies. However, comparing with the Transformer Base, the improvement of the Transformer Big on long sentences ($+1.20$ BLEU for sentences with more than $45$ tokens) was similar to that on short sentences ($+1.14$ BLEU for sentences with no more than $15$ tokens), while our approach to model phrases in the Transformer model even brings significantly ($p<0.01$) more improvements ($+1.72$ BLEU) on the performance of longer sentences with the Transformer Base setting (8 heads) than the Transformer Big with $16$ heads ($+1.20$ BLEU).

The length analysis result is consistent with our conjecture to some extent given that there are likely to be more long-distance dependencies in longer source sentences. We suggest that phrase sequences which are shorter than corresponding token sequences can help the model capture long-distance dependencies better, and modeling phrase representations for the Transformer can enhance its performance on long sequences.

\begin{figure}[t]
\centering
\includegraphics[width=1.0\columnwidth]{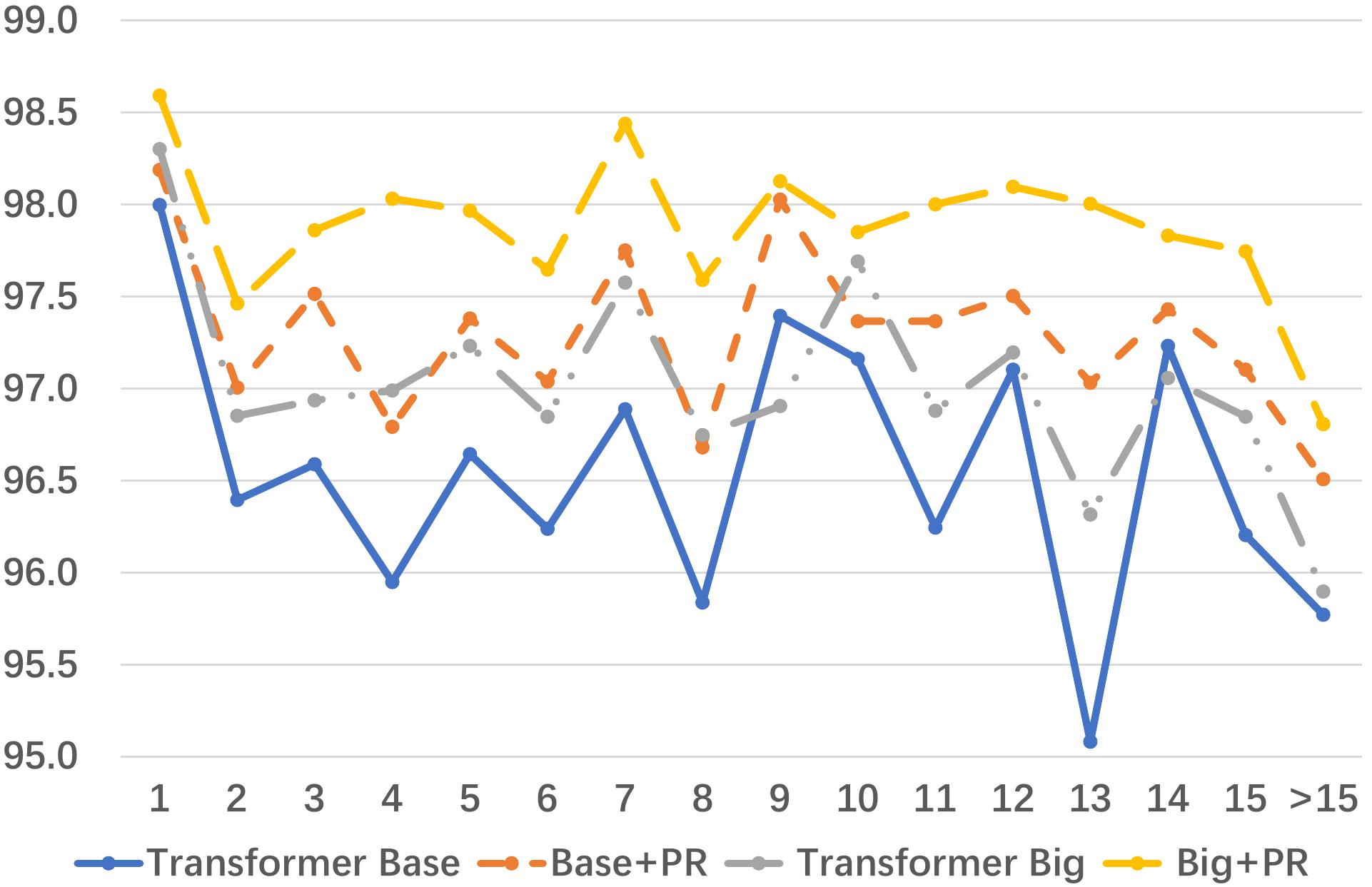}
\caption{Subject-Verb Agreement Analysis. X-axis and y-axis represent subject-verb distance in words and the accuracy respectively.}
\label{ling}
\end{figure}

\subsection{Subject-Verb Agreement Analysis}

Intuitively, in translating longer sentences we should encounter more long-distance dependencies than in short sentences. To verify whether our method can improve the capability of the NMT model to capture long-distance dependencies, we also conducted a linguistically-informed verb-subject agreement analysis on the \emph{Lingeval97} dataset \cite{sennrich2017grammatical} following \newcite{tang2018self}.

In German, subjects and verbs must agree with one another in grammatical number and person. In \emph{Lingeval97}, each contrastive translation pair consists of a correct reference translation, and a contrastive example that has been minimally modified to introduce one translation error. The accuracy of a model is the number of times it assigns a higher score to the reference translation than to the contrastive one, relative to the total number of predictions. Results are shown in Figure \ref{ling}.

Figure \ref{ling} shows that our approach can improve the accuracy of long-distance subject-verb dependencies, especially for cases where there are more than $10$ tokens between the verb and the corresponding subject when comparing the ``Base+PR'' with the ``Transformer Big''.

\section{Conclusion}

Considering that the strong Transformer translation model still has difficulty in fully capturing long-distance dependencies \cite{tang2018self}, and that using a shorter phrase sequence (in addition to the original token sequence) is an intuitive approach to help the model capture long-distance features, in this paper, we first propose an attention mechanism to generate phrase representations by merging corresponding token representations. In addition, we incorporate the generated phrase representations into the Transformer translation model to help it capture long-distance relationships. We obtained statistically significant improvements on the WMT 14 English-German and English-French tasks over the strong Transformer baseline, which demonstrates the effectiveness of our approach. Our further analysis shows that the Transformer with phrase representation empirically improves its performance especially in long-distance dependency learning.

\section*{Acknowledgments}

We thank anonymous reviewers for their insightful comments and helpful advice. Hongfei Xu acknowledges the support of China Scholarship Council ([2018]3101, 201807040056). Deyi Xiong is supported by the National Natural Science Foundation of China (Grant No. 61861130364), the Natural Science Foundation of Tianjin (Grant No. 19JCZDJC31400) and the Royal Society (London) (NAF$\backslash$R1$\backslash$180122). Hongfei Xu, Josef van Genabith and Jingyi Zhang are supported by the German Federal Ministry of Education and Research (BMBF) under the funding code 01IW17001 (Deeplee).

\bibliography{acl2020}
\bibliographystyle{acl_natbib}

\end{document}